\documentclass[letterpaper]{article} 
\usepackage{aaai25}  
\usepackage{times}  
\usepackage{helvet}  
\usepackage{courier}  
\usepackage[hyphens]{url}  
\usepackage{graphicx} 
\usepackage{natbib}  
\usepackage{caption} 
\usepackage{algorithm}
\usepackage{algorithmic}

\usepackage{subcaption}
\usepackage{mathrsfs}
\usepackage{multirow}
\usepackage{amsmath}
\usepackage{booktabs}
\usepackage{xcolor}

\usepackage{newfloat}
\usepackage{listings}
\DeclareCaptionStyle{ruled}{labelfont=normalfont,labelsep=colon,strut=off} 
\floatstyle{ruled}
\newfloat{listing}{tb}{lst}{}
\floatname{listing}{Listing}
\title{GADT: Enhancing Transferable Adversarial Attacks through Gradient-guided Adversarial Data Transformation}
\author{
    Yating Ma,
    Xiaogang Xu,
    Liming Fang,
    Zhe Liu
}
\usepackage{bibentry}

\begin{document}

\maketitle

\begin{abstract}
Current Transferable Adversarial Examples (TAE) are primarily generated by adding Adversarial Noise (AN). Recent studies emphasize the importance of optimizing Data Augmentation (DA) parameters along with AN, which poses a greater threat to real-world AI applications. However, existing DA-based strategies often struggle to find optimal solutions due to the challenging DA search procedure without proper guidance.
In this work, we propose a novel DA-based attack algorithm, \textbf{GADT}. GADT identifies suitable DA parameters through iterative antagonism and uses posterior estimates to update AN based on these parameters. We uniquely employ a differentiable DA operation library to identify adversarial DA parameters and introduce a new loss function as a metric during DA optimization. This loss term enhances adversarial effects while preserving the original image content, maintaining attack crypticity.
Extensive experiments on public datasets with various networks demonstrate that GADT can be integrated with existing transferable attack methods, updating their DA parameters effectively while retaining their AN formulation strategies.
Furthermore, GADT can be utilized in other black-box attack scenarios, e.g., query-based attacks, offering a new avenue to enhance attacks on real-world AI applications in both research and industrial contexts.
\end{abstract}

%

\section{Introduction}
\label{sec:intro}

Artificial Intelligence (AI) has made tremendous progress with Deep Neural Networks (DNNs) in various high-security tasks, such as face recognition \citep{meng2021magface, boutros2022elasticface}, disease diagnosis \citep{khan2021machine}, and recent large-scale vision-language deep models \citep{radford2021learning, li2023semantic}.
However, all existing DNNs still suffer from the safety issue of Adversarial Examples (AEs), which can cause the target model to give incorrect results by adding human-imperceptible noise.
Among various attack methods, Transferable Attacks(TAs) have garnered significant attention because they require minimal knowledge of the target model, fitting into the practical category of black-box attacks.
In these scenarios, attackers use a white-box surrogate model to generate AEs and evaluate the attack success rate on the target model.
By studying TAs, researchers and developers can train more robust and reliable models, enhancing their security and trustworthiness. Therefore, improving the transferability of AEs is an urgent topic that needs to be explored.

\begin{figure}[t]
    \centering
    \begin{minipage}[t]{0.45\columnwidth}
		\centering
		\centerline{\includegraphics[width=\columnwidth]{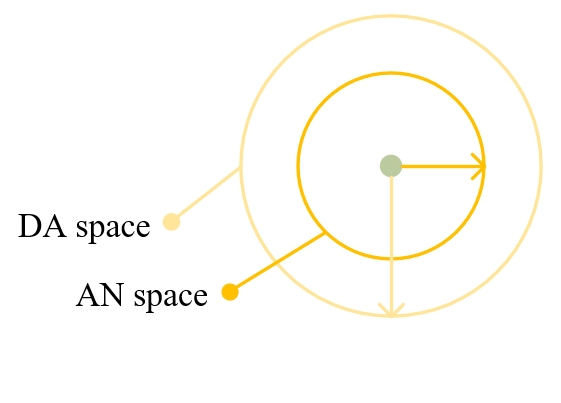}}
		\subcaption{Solution Space (S.S.)}
		\label{figs:solution_1}
    \end{minipage} 
    \begin{minipage}[t]{0.45\columnwidth}
		\centering
		\centerline{\includegraphics[width=\columnwidth]{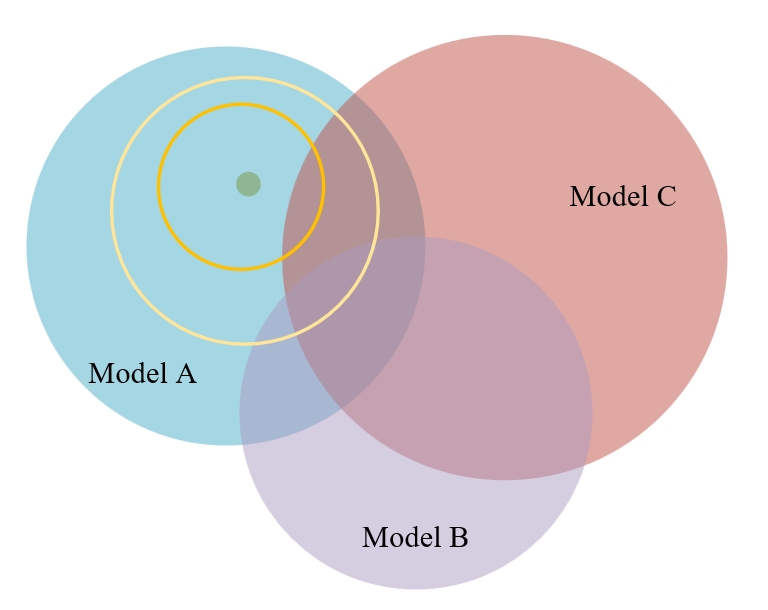}}
		\subcaption{S.S. v.s. Attack Space }
		\label{figs:solution_2}
    \end{minipage} 
    \vspace{-0.1in}
    \caption{
    This illustration shows the solution space for AN- and DA-based methods. 
    The DA-based strategy typically offers a larger solution space by incorporating transformations in addition to noise (a). This expanded solution space can more effectively encompass the attack samples' space across various target models.
    }
    \label{figs:solution}
    \vspace{-0.2in}
\end{figure}

The algorithm optimization for TAs can be divided into two categories: searching for better Adversarial Noises (AN) using gradient-related information and adjusting Data Augmentation (DA) parameters. The former is termed the ``AN-based" method, while the latter is known as the ``DA-based" strategy. Representative AN-based approaches~\citep{han2023sampling, yang2023improving, zhou2018transferable, fang2024strong}, such as MI-FGSM~\citep{dong2018boosting} and NI-FGSM~\citep{lin2019nesterov}, primarily design variants of gradients extracted from the surrogate model. These variants can be obtained by employing different losses~\citep{xiong2022stochastic}, ensembling gradients from different iterations~\citep{dong2018boosting}, and so on.
However, nearly all existing AN-based methods exhibit low transferability in black-box settings when target models' architectures differ from surrogate models. 
This limitation arises because the attack solution space with AN is insufficient for adapting to diverse target models, as shown in Fig.~\ref{figs:solution}. As \citet{liu2016delving} pointed out, the decision boundaries of different models highly overlap. DA-based methods expand the generation space of adversarial examples, increasing the likelihood of finding these overlapping regions (Fig.~\ref{figs:solution_2}).

By contrast, DA-based methods~\citep{dong2019evading, lin2019nesterov, xie2019improving, lin2024boosting} involve adjusting DA parameters along with AN. They focus on optimizing the combination of AN and various DA operations. For instance, DI-FGSM~\citep{xie2019improving} uses random transformation operations along with AN generated by FGSM. This suggests that DA operations can diversify the data, expand the generation space of adversarial examples, and reduce dependency on the surrogate model. However, these strategies have not optimized DA parameters and may cause suboptimal results. 
Thus, subsequent solutions have proposed an additional search procedure. For example, \citet{yan2022ila} introduced ILA-DA, an automated strategy focused on finding the optimal combination of pre-defined transformations. However, in existing search-based methods, DA parameters obtained are often still suboptimal due to the lack of direct computation of the gradient relationship between the attack metric and augmentation parameters.

In this work, we propose a novel DA-based attack which directly optimizes DA operations using the raw gradient information of the attack metric concerning DA parameters. This strategy, called GADT, can be combined with any TAs, as its core principle lies in a new DA optimization paradigm (including these DA-based methods, since we can apply our strategy to optimize their DA parameters after their DA search or sampling procedure is completed).

We propose a novel method for updating DA parameters. Unlike existing methods that search for different combinations of DA parameters, we update DA parameters directly based on the gradient direction of the attack metric with respect to the DA parameters. We employ differentiable DA operations to compute the corresponding raw gradients, using Kornia~\citep{riba2020kornia}, a differentiable computer vision library that includes data augmentation operators. Although the range of differentiable DA operations is limited, they yield better attack effects compared to traditional combinations of more DA operations. Also, the acquisition cost of such optimal DA parameters are lower than traditional strategies that utilize heavy search procedures.

Furthermore, we design a new loss function that serves as the attack metric for updating DA parameters. 
Despite of its simpleness, it can guide the DA optimization in a satisfied manner.
The main advantage of this metric is its ability to simultaneously identify the optimal attack solution for DA parameters while preserving the original image content. This enhances the stealthiness of adversarial examples, making them harder to detect and thereby increasing their threat.

We conducted extensive experiments on public datasets. The results demonstrate our approach's effectiveness in improving the performance of TAs across different networks.
In summary, our main contributions are three-fold:
\begin{itemize}
    \item[$ \bullet $] We propose a novel attack-oriented strategy to formulate offensive DA operations, utilizing the raw gradient data of the attack metric with respect to DA parameters. 
    \item[$ \bullet $] We design a new loss function as the attack metric that considers both aggressivity and crypticity.
    \item[$ \bullet $] We conducted extensive evaluations on public datasets using various networks and baselines, demonstrating that our strategy achieves stronger Transferable Attacks (TAs). Additionally, our strategy has proven effective for other black-box attacks, e.g., query-based attacks.
\end{itemize}

\begin{figure*}[!ht]
    \centering
    \includegraphics[width=0.8\linewidth]{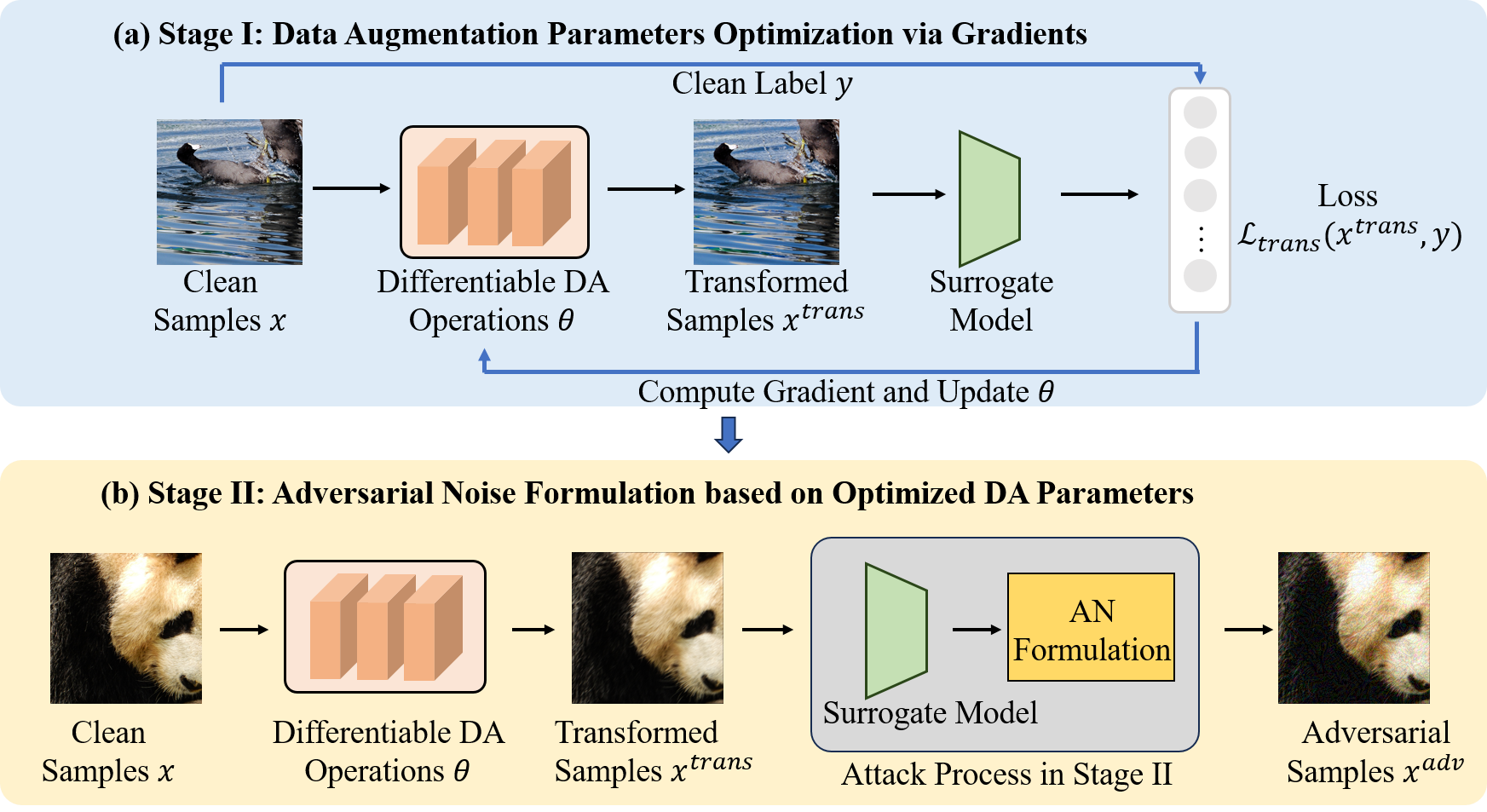}
    \caption{
    The pipeline of our attack method. We begin by identifying optimal DA parameters that induce adversarial effects, leveraging gradient information from DA operations and a novel loss function for guidance. This automated transformation procedure can then be integrated with any AN-based attack process to generate highly effective transferable adversarial examples.
    }
    \label{figs:process}
    \vspace{-0.2in}
\end{figure*}

\section{Related Work}
\label{sec:related}


Current research on transferable attacks can be broadly categorized into two main approaches: gradient-optimization-based methods \citep{han2023sampling, yang2023improving, wan2023adversarial, zhu2023boosting} and DA-based methods \citep{dong2019evading, lin2019nesterov, xie2019improving, lin2024boosting}. We will provide a brief overview of both, with a closer focus on the latter, as it is more closely related to our work.

\textbf{Gradient-optimization-based methods.} 
These approaches were initially developed for white-box attacks to improve gradients for formulating AN. However, they have also proven effective for transferable attacks in black-box scenarios.
MI-FGSM~\citep{dong2018boosting}, based on FGSM~\citep{goodfellow2014explaining}, is a significant method in transferable attacks that introduces momentum during the generation of adversarial examples.
This approach accelerates the gradient descent process and enhances the attack success rate.
Later, the Nesterov accelerated gradient method \citep{lin2019nesterov} was introduced as an optimization algorithm for minimizing convex functions, incorporating momentum to accelerate convergence.
Subsequently, \citet{wang2021enhancing} utilized gradient variance from previous iterations to adjust the current gradient, stabilizing the update direction and avoiding poor local optima.
However, these strategies still exhibit limited transferability in black-box settings because the effective attack spaces for different target models are broad, and relying solely on AN is not sufficient to cover them.

\textbf{Data-augmentation-based methods.} 
DA-based approaches transform clean examples using various combinations of DA parameters and then input them into surrogate models to compute gradients and generate adversarial examples.
DI-FGSM \citep{xie2019improving} introduces random perturbations, including color and texture variations, at each iteration to enhance the robustness of adversarial examples.
TI-FGSM \citep{dong2019evading} adopts a translation-invariant approach to correct the gradient direction, achieved through predefined kernel convolutions for image translation.
SI-NI-FGSM \citep{lin2019nesterov} utilizes Nesterov momentum to escape local optima during optimization, leveraging the scale-invariance property of DNNs to enhance transferability.
\citet{wang2021admix} proposed Admix, a novel input transformation that blends the original image with randomly selected images from other classes through linear interpolation, calculating gradients for the blended image while preserving the original label.
However, none of these methods have addressed the optimization of data augmentation parameters.

Thus, a series of search-based frameworks have been proposed. ILA-DA \citep{yan2022ila} employs three novel augmentation techniques to improve adversarial examples by maximizing their perturbation on an intermediate layer of the surrogate model. It focuses on finding the optimal combination weight for each augmentation operation.
ATTA \citep{wu2021improving}, which is related to our method, constructs a DNN to simulate data augmentation but only considers color and texture variations. Similar to ILA-DA, it focuses on optimizing the combination weight. However, these methods are limited in finding optimal DA parameters because they mainly consider the combination of different DA operations without optimizing the parameters specific to each operation. Moreover, they lack direct guidance on optimizing each operation individually.





\section{Method}
\subsection{Preliminary}
\label{sec:problem}

\noindent\textbf{The formulation of adversarial attack.} Let's consider a image classification model $f(x)$, where $x$ denotes the input and $y$ is the corresponding ground truth. The attacker generates an adversarial example as $x^{adv}=x+\delta $, where $\delta$ is the designed perturbation, and $\delta$ is restrained by $l_{p } $-ball.
For the adversarial sample $x^{adv}$, its output should satisfy $f(x^{adv})\neq y$.
The traditional pipeline for generating such adversarial samples involves an standard optimization problem, which can be formulated as follows
\begin{equation}
    \begin{aligned}
        &arg\underset{x^{adv}}{\max} \quad \mathcal{L}(x^{adv},y),   \\
        &s.t. \quad \left \| x^{adv}-x \right \|_\infty \le \varepsilon
    \end{aligned}
    \label{eq:1}
\end{equation}
where $\mathcal{L}(\cdot,\cdot )$ is the loss function, $\varepsilon$ is the maximum perturbation range on the $l_{\infty } $-ball.
To solve the optimization problem in Eq.~\ref{eq:1}, an iterative method is typically employed. In this approach, adversarial examples are generated based on the gradient direction of the loss function, as follows
\begin{equation}
    \begin{split}
    &x_0^{adv}=x,\\
        x_{t+1}^{adv}=x_{t}^{adv}+ \alpha &\times sign(\bigtriangledown _{x_t^{adv}}\mathcal{L}(f(x_t^{adv}),y) ), 
    \end{split}
    \label{eq:2}
\end{equation}
where $t$ denotes the $t\text{-}th$ iteration, $\alpha$ represents the step size, $sign(\cdot)$ is the sign function.

\noindent\textbf{Augmentation can help enhance attack effects.}
Several methods utilize augmentation strategies to enhance attack efficacy. The fundamental theory is that suitable augmentation can expand the solution space of adversarial examples, thereby increasing the attack success rate (as shown in Fig.~\ref{figs:solution}). In contrast to traditional adversarial examples, which are generated by simply adding noise, augmentation incorporates various transformations (e.g., spatial, color changes) to the original clean samples.

In this approach, the adversarial sample is first augmented through various operations and then finalized by formulating the adversarial perturbations using the attack pipeline, as in Eq.~\ref{eq:2}. In our work, we employ a parameterizable augmentation module for the augmentation step. Optimizing the corresponding parameters allows us to obtain aggressive DA parameters that are suitable for the attack.
Suppose the augmentation can be formulated as follows
\begin{equation}
    x^{trans}=\mathcal{T}_{\theta}(x),
    \label{eq:3}
\end{equation}
where $\mathcal{T}$ is the augmentation module with $\theta$ as its corresponding parameter. While $\mathcal{T}$  can be a neural network, it typically lacks interpretability and generalization, and its parameter $\theta$ is often large, making it unsuitable for many applications. In this work, we use a differentiable data augmentation library that provides interpretable augmentation operations and adjustable parameters $\theta$, which are tiny and more suitable for our needs.

\subsection{Motivation}
\label{sec:motivation}
\noindent\textbf{The shortcomings of existing DA-based attack methods.}
Although there have been several DA-based attack methods, they come with various disadvantages. As discussed in ``Introduction Section", existing DA-based attacks can be classified into two categories. One approach involves using traditional transformation strategies and exploring combinations of different DA operations to achieve better attack results. However, the number of possible combinations of DA operations is vast, and different models may require varying combinations, which adds to the complexity. For example, simpler models like VGG16\citep{simonyan2014very} may achieve successful attacks with small magnitude transformations, while more complex models like Inception-v3\citep{szegedy2016rethinking} may require more extensive transformations. Therefore, relying on empirically determined or randomly sampled DA parameters often leads to unsatisfactory results.
To address the challenge of empirical combination, another approach involves automated search strategies\citep{wu2021improving, yan2022ila}, which aim to find optimal combinations of transformations. However, these methods lack a direct optimization direction for DA parameters and heavily rely on final classification results, making them highly ill-posed. In summary, existing DA-based approaches lack a straightforward optimization direction for DA parameters, often resulting in suboptimal parameter choices.

\noindent\textbf{Our strategy with gradient-guided optimization direction for DA parameters.}
Given the existing challenges analyzed above, our goal is to enhance the diversity of augmentation and steer data transformations towards optimizing parameters that significantly benefit the attack objective. To achieve this, we use the loss function as a guide for optimization. We update the transformation parameters in the direction of gradient increase (the gradient of loss towards DA parameters), akin to common attack methodologies. Moreover, we iterate this process multiple times to expand the solution space of adversarial examples and identify the optimal transformation parameters.

Typically, DA parameters are non-differentiable. However, there are now differentiable DA libraries such as Kornia~\citep{riba2020kornia} that implement operations with adjustable parameters, as described in Eq.~\ref{eq:3}. Although these libraries support only a subset of DA operations, they can significantly aid in achieving attack objectives by providing direct and accurate gradient guidance. This stands in contrast to traditional methods that rely on sub-optimal combinations of varied DA parameters.

Our attack method consists of two main steps, illustrated in Figure~\ref{figs:process}. First, we expand the solution space and perform an automated search for optimal DA parameters, guided by gradient information. Second, we input the transformed images into existing attack processes such as MI-FGSM~\citep{dong2018boosting} to generate final attack results. Importantly, our method can seamlessly integrate into any existing attack strategy capable of producing effective adversarial perturbations, thanks to its independence (including these DA-based strategies, since our approach can be applied to optimize DA parameters obtained through their search process as well).

\subsection{Our Implementation}
\label{sec:network}

In this study, we integrate two transformation techniques—motion blur and saturation adjustment—into our data augmentation module denoted as $\mathcal{D}$ using Kornia~\citep{riba2020kornia}. Herein, we detail the procedure for generating the aggressive data augmentation parameters, the key is the gradient-based guidance and the new loss function.

\noindent\textbf{Data Augmentation based on Kornia.} 
Kornia is a comprehensive computer vision library comprising modules with operators designed for seamless integration into neural networks. Built on PyTorch, Kornia enables reverse-mode auto-differentiation to compute gradients of augmentation transformations. This capability optimizes data transformations during training, similar to model training itself. With Kornia, we achieve precise control over augmentation parameters, facilitating gradient computation of the loss function with respect to each transformation's magnitude effortlessly.

\begin{algorithm}[t]
\caption{GADT}\label{alg:attack}
\textbf{Input}: A clean image $x$ and its ground-truth label $y$, transformation network $\mathcal{T}(\cdot)$ and its paramethers $\theta$, loss function $\mathcal{L}_{trans}$, number of transformation iterations $K$, surrogate model $f(\cdot)$\\
\textbf{Parameter}: Perturbation budget $\varepsilon$, number of attack iteration $T$, classify loss function $\mathcal{L}$, momentum $\mu$\\
\textbf{Output}: $\theta, x^{adv}$
\begin{algorithmic}[1]
\STATE Initialize $x_0^{trans}=x$, $\theta_{k-1}$
\FOR {$k=0$ to $K-1$}
\STATE $x_k^{trans}=\mathcal{T}(x_{k}^{trans};\theta)$
\STATE Update $\theta_k = \theta_k - Adam(\mathcal{L}_{trans}(x_k^{trans},y))$
\ENDFOR
\STATE $\alpha=\varepsilon/T$; $g_0=0$; $x_0^{adv}=x_K^{trans}$ 
\FOR {$t=0$ to $T-1$}
\STATE Input $x_t^{adv}$ to obtain the gradient $\bigtriangledown_x\mathcal{L}(x_t^{adv},y)$
\STATE Update $g_{t+1}=\mu \cdot g_t + \frac{\bigtriangledown _{x}\mathcal{L}(x_t^{adv},y)}{\parallel\bigtriangledown _{x}\mathcal{L}(x_t^{adv},y) \parallel _{1} } $
\STATE Update $x_{t+1}^{adv}=x_t^{adv}+\alpha \cdot sign(g_{t+1})$
\ENDFOR 
\STATE \textbf{return} $x_T^{adv}$
\end{algorithmic}
\end{algorithm}

\noindent\textbf{The new loss function to guide the DA parameters' optimization.}
To determine the optimal DA parameters, we iteratively apply data augmentation to each clean sample, adjusting transformation parameters based on gradient ascent. A critical aspect is selecting a suitable loss function to guide this process. While a straightforward approach involves using task-oriented losses like Cross-Entropy (CE) for classification tasks, we must also consider adversarial stealthiness. Excessive augmentation can not only enhance attack efficacy but also risks detection by intelligent systems. 
Our goal is to devise a new loss function that serves as the metric, balancing the maximization of attack efficacy with the preservation of original image content, thereby achieving both objectives simultaneously.

Specifically, for the attack target, we employ the CE loss, denoted as $\mathcal{L}_{CE}$. Additionally, we utilize the Mean Squared Error (MSE) loss, $\mathcal{L}_{MSE}$, to enforce fidelity at the pixel level between adversarial examples and clean samples. 
To integrate these objectives, we combine both loss functions with a balancing parameter $\lambda$. The overall loss function is formulated as follows
\begin{equation}
\small
    \begin{split}
        \mathcal{L}_{trans} = -\mathcal{L}_{CE}(f(x^{trans}),y) + \lambda \cdot \mathcal{L}_{MSE}(x,x^{trans}).
    \end{split}
    \label{eq:loss}
\end{equation}
Despite the simplicity of this loss function, we have found it to be highly effective in DA-based attacks, ensuring both adversarial potency and the crypticity of AEs.

\noindent\textbf{Iterative update for optimal DA parameters.}
We update the DA parameters $\theta$ based on the loss function in Eq.~\ref{eq:loss}. Drawing inspiration from adversarial attacks, we optimize the transformation parameters iteratively to enhance their adversarial effects, as follows
\begin{equation}
    \begin{split}
        \theta_{k+1} = \theta_k - Adam_{\theta_k}(\mathcal{L}_{trans}),
    \end{split}
\end{equation}
where $k$ denote the $k\text{-}th$ iteration, and $Adam_{\theta_k}(\mathcal{L}_{trans})$ is the update computed by backpropagating the gradient of $\mathcal{L}_{trans}$ towards $\theta_k$.
After iterative optimization of the DA parameters, the adversarial perturbation can be formulated based on the augmented data. The overall attack procedure, which integrates our DA strategy with MI-FGSM, is summarized in Algorithm~\ref{alg:attack}.

\section{Experiments}
\label{sec:experiment}


\renewcommand{\dblfloatpagefraction}{.9}
\begin{table*}[h!]
    \centering
    \small
    \renewcommand{\arraystretch}{1.05}
    \setlength{\tabcolsep}{0.5mm} {\begin{subtable}[h!]{1.0\linewidth}
        \centering
        \begin{tabular}{c|c|c|c|c|c|c|c|c|c}
        \hline
        \small
        Surrogate & Attack & VGG16 & \makebox[0.08\textwidth][c]{RN50} & RN101 & \makebox[0.08\textwidth][c]{Inc-v3} & DN121 & IncRes-v2 & $\rm CLIP_{RN101}$ & $\rm CLIP_{ViT/B32}$\\ \hline
        \multirow{2}{*}{VGG16} 
        ~ & MIM & 98.4  & 78.0  & 65.6  & 60.6  & 74.4  & 48.8  & 71.7 & 36.8 \\ 
        ~ & GADT-MIM & \textbf{99.7}  & \textbf{93.3}  & \textbf{85.3}  & \textbf{82.6}  & \textbf{90.6}  & \textbf{74.2}  & \textbf{89.7} & \textbf{52.5}  \\ \hline
        \multirow{2}{*}{RN101} 
        ~ & MIM & 84.5  & 95.8  & 98.2  & 58.3  & 82.4  & 47.9  & 61.7 & 40.4  \\ 
        ~ & GADT-MIM & \textbf{96.2}  & \textbf{98.6}  & \textbf{98.9}  & \textbf{80.0}  & \textbf{96.5}  & \textbf{73.0}  & \textbf{84.5} & \textbf{63.2}   \\ \hline
        \multirow{2}{*}{Inc-v3} 
        ~ & MIM & 74.3  & 60.6  & 50.2  & 98.6  & 54.4  & 55.0  & 53.2 & 30.6  \\ 
        ~ & GADT-MIM & \textbf{89.2}  & \textbf{79.0}  & \textbf{70.3}  & \textbf{99.0}  & \textbf{76.7}  & \textbf{77.1}  & \textbf{70.6} & \textbf{42.0}  \\  \hline
        \multirow{2}{*}{DN121} 
        ~ & MIM & 90.0  & 93.4  & 88.1  & 73.2  & 97.3  & 60.1  & 69.3 & 46.9  \\ 
        ~ & GADT-MIM & \textbf{98.4}  & \textbf{98.0}  & \textbf{95.4}  & \textbf{89.9}  & \textbf{98.8}  & \textbf{82.1}  & \textbf{88.0} & \textbf{72.1}  \\ \hline
        \end{tabular}
        \vspace{-0.05in}
        \caption{The transferable attack results when combine our method with MIM.}
        \vspace{0.05in}
    \end{subtable}}
    \setlength{\tabcolsep}{0.5mm}{\begin{subtable}[h!]{1.0\linewidth}
    \centering
    \small
    \renewcommand{\arraystretch}{1.05}
        \begin{tabular}{c|c|c|c|c|c|c|c|c|c}
        \hline
        Surrogate & Attack & VGG16 & \makebox[0.08\textwidth][c]{RN50} & RN101 & \makebox[0.08\textwidth][c]{Inc-v3} & DN121 & IncRes-v2 & $\rm CLIP_{RN101}$ & $\rm CLIP_{ViT/B32}$ \\ \hline
        \multirow{2}{*}{VGG16} 
        ~ & SIM & 92.9  & 55.9  & 43.5  & 46.5  & 50.1  & 33.5  & 54.5 & 54.5 \\ 
        ~ & GADT-SIM & \textbf{98.3}  & \textbf{81.6}  & \textbf{67.9}  & \textbf{69.7}  & \textbf{77.8}  & \textbf{60.2}  & \textbf{76.3} & \textbf{46.6} \\ \hline
        \multirow{2}{*}{RN101} 
        ~ & SIM & 71.4  & 69.2  & 86.0  & 51.5  & 60.2  & 38.7  & 51.8 & 61.1  \\ 
        ~ & GADT-SIM & \textbf{93.1}  & \textbf{89.6}  & \textbf{92.1}  & \textbf{73.4}  & \textbf{82.8}  & \textbf{65.7}  & \textbf{73.3} & \textbf{52.7}   \\ \hline
        \multirow{2}{*}{Inc-v3} 
        ~ & SIM & 50.5  & 38.6  & 33.0  & 70.4  & 37.6  & 28.3  & 36.5 & 46.5 \\ 
        ~ & GADT-SIM & \textbf{73.6}  & \textbf{60.6}  & \textbf{53.3}  & \textbf{80.0}  & \textbf{59.9}  & \textbf{48.1}  & \textbf{52.5} & \textbf{36.4}  \\  \hline
        \multirow{2}{*}{DN121} 
        ~ & SIM & 74.4  & 62.3  & 55.2  & 47.2  & 86.1  & 35.7  & 54.6 & 63.8 \\ 
        ~ & GADT-SIM & \textbf{93.0}  & \textbf{85.1}  & \textbf{78.7}  & \textbf{70.5}  & \textbf{94.4}  & \textbf{63.5}  & \textbf{77.8} & \textbf{54.6} \\  \hline
        \end{tabular}
        \vspace{-0.05in}
        \caption{The transferable attack results when combine our method with SIM.}
        \vspace{0.05in}
    \end{subtable}}
    
    \setlength{\tabcolsep}{0.5mm}{\begin{subtable}[h!]{1.0\linewidth}
    \centering
    \small
    \renewcommand{\arraystretch}{1.05}
        \begin{tabular}{c|c|c|c|c|c|c|c|c|c}
        \hline
        Surrogate & Attack & VGG16 & \makebox[0.08\textwidth][c]{RN50} & RN101 & \makebox[0.08\textwidth][c]{Inc-v3} & DN121 & IncRes-v2 & $\rm CLIP_{RN101}$ & $\rm CLIP_{ViT/B32}$ \\ \hline
        \multirow{2}{*}{VGG16} 
        ~ & Admix & 97.9  & 76.3  & 65.3  & 69.4  & 77.2  & 54.1  & 66.8 & 54.5 \\ 
        ~ & GADT-Admix & \textbf{99.3}  & \textbf{91.7}  & \textbf{84.8}  & \textbf{85.3}  & \textbf{92.3}  & \textbf{75.9}  & \textbf{82.4} & \textbf{63.8}    \\ \hline
        \multirow{2}{*}{RN101} 
        ~ & Admix & 80.7  & 85.6  & 93.9  & 70.2  & 82.2  & 62.0 & 61.6 & 61.1 \\ 
        ~ & GADT-Admix & \textbf{94.0}  & \textbf{93.5}  & \textbf{96.6}  & \textbf{86.4}  & \textbf{93.0}  & \textbf{79.0}  & \textbf{81.6} & \textbf{67.1}   \\ \hline
        \multirow{2}{*}{Inc-v3} 
        ~ & Admix & 73.3  & 67.6  & 61.3  & 91.1  & 69.5  & 58.7  & 55.4 & 46.5 \\ 
        ~ & GADT-Admix & \textbf{88.5}  & \textbf{83.3}  & \textbf{79.3}  & \textbf{95.2}  & \textbf{83.9}  & \textbf{78.5}  & \textbf{72.5} & \textbf{55.6} \\ \hline
        \multirow{2}{*}{DN121} 
        ~ & Admix & 85.5  & 81.9  & 78.7  & 71.2  & 96.1  & 59.6  & 65.1 & 63.8  \\ 
        ~ & GADT-Admix & \textbf{95.6}  & \textbf{93.5}  & \textbf{90.8}  & \textbf{86.5}  & \textbf{98.1}  & \textbf{79.6}  & \textbf{84.2} & \textbf{69.3}  \\  \hline
        \end{tabular}
        \caption{The transferable attack results when combine our method with Admix.}
        \label{subtab:admix}
    \end{subtable}}
    \vspace{-0.05in}
    \caption{The attack evaluation when combine our proposed GADT with different attack methods. Equipped with our GADT, the performance of all attack methods can be improved.}
    \label{tab:attack}
    \vspace{0.05in}
\end{table*}

\subsection{Experimental Settings}

\noindent\textbf{Dataset.}
We conducted experiments on a dataset of 1,000 images extracted from an ImageNet-compatible dataset, used in the NIPS 2017 adversarial competition\footnote{https://www.kaggle.com/datasets/google-brain/nips-2017-adversarial-learning-development-set}. This dataset is widely utilized for evaluating transferable attack methods~\citep{kurakin2018adversarial}.

\noindent\textbf{Models.}
We selected five commonly used undefended models as surrogate models: VGG16~\citep{simonyan2014very}, ResNet-101 (RN101)~\citep{he2016deep}, Inception v3 (Inc-v3)~\citep{szegedy2016rethinking}, and DenseNet-121 (DN121)~\citep{huang2017densely}. These models were also employed as target models for evaluation. Additionally, to comprehensively assess attack effects in the black-box setting, we included ResNet-50 (RN50)~\citep{he2016deep}, Inception-ResNet v2 (IncRes-v2)~\citep{szegedy2017inception}, and CLIP (ResNet-101 
\& ViT-B/32 version)~\citep{radford2021learning}, which is a prominent vision-language model, alongside the aforementioned models.

For evaluating adversarial defense methods, we consider the adversarially trained Inception-v3 model ($\rm {Inc\text{-}v3}_{adv}$)~\citep{kurakin2018adversarialattacksdefencescompetition}, as well as two methods: AT~\citep{tramer2017ensemble} and HGD~\citep{liao2018defense}.
All these models are pretrained on the ImageNet's valuation set.

\noindent\textbf{Baselines.}
We compare our method with various state-of-the-art transferable transformation-based attack methods mentioned in related work: Momentum Iterative Fast Gradient Sign Method (MIM)~\citep{dong2018boosting}, Diverse Input Method (DIM)~\citep{xie2019improving}, Translation-Invariant Method (TIM)~\citep{dong2019evading}, ScaleInvariant Method (SIM)~\citep{lin2019nesterov}, and Admix~\citep{wang2021admix}.
Specifically, our strategy focuses on optimizing the DA parameters of these methods to evaluate improvements in attack effectiveness. Among them, DIM, TIM, and SIM are all transfer attack methods that rely on input transformation. In our experiments, ``GADT-X" means applying our GADT strategy on the baseline of X.

\noindent\textbf{Attack details.}
For all attack methods, we followed the parameter settings used in the original papers. 
We set $\alpha =1.6$, the number of attacks $T=10$, and the perturbation size $\varepsilon =16/255$. For our method, we set $\lambda =1$, and the initial values of motion blur and saturation to 0.5 and (0.75, 0.75), respectively. The number of iterations for GADT is 20. 

\subsection{Transferable Attack Results}
As shown in Table~\ref{tab:attack}, we incorporate our method into various attack approaches targeting four surrogate models to produce adversarial examples using $\rm GADT$. Compared to baseline methods, $\rm GADT$ demonstrates superior performance. For instance, when combined with MIM, the transferable success rate increases by over 15\%. This highlights the effectiveness of optimizing DA parameters using the gradient guidance.
Additionally, our method can be combined with previous DA-based methods like DIM and TIM to enhance attack capabilities(See in Appendix A). Applying our DA optimization after their DA search phase results in a 5\% to 20\% increase in attack success rates compared to the corresponding baselines.

Additionally, we tested the attack effectiveness on two versions of the visual-language model CLIP, such as $\rm CLIP_{RN101}$ and $\rm CLIP_{ViT-B/32}$. 
The experimental results demonstrate that our method exhibits higher transferability of attacks on both the CNN and ViT architectures. For both models, the improvement in attack success rate is generally above 10\%.

\begin{table}[h!]
\small
\centering
\vspace{-0.1in}
\renewcommand{\arraystretch}{1.05}
\setlength{\tabcolsep}{0.65mm}{\begin{tabular}{lllll}
\hline
Surrogate               & Attack   & $\rm Inc\text{-}v3_{adv}$ & AT   & HGD  \\ \hline
\multirow{6}{*}{Inc-v3} & MIM      & 55.0    & 47.2 & 1.3  \\
                        ~ & SIM      & 28.3    & 47.8 & 1.2  \\
                        ~ & DIM      & 70.2    & 47.7 & 1.7  \\
                        ~ & TIM      & 66.9    & 47.4 & 3.9  \\
                        ~ & Admix    & 58.7    & 50.3 & 5.0  \\ \cline{2-5} 
                        ~ & GADT-TIM & \textbf{84.8}    & \textbf{51.2} & \textbf{8.3}  \\ \hline
\multirow{6}{*}{DN121}  & MIM      & 60.1    & 47.4 & 40.3 \\
                        ~ & SIM      & 35.7    & 47.5 & 20.2 \\
                        ~ & DIM      & 74.7    & 48.1 & 60.0 \\
                        ~ & TIM      & 71.7    & 48.9 & 78.4 \\
                        ~ & Admix    & 59.6    & 49.7 & 50.8 \\ \cline{2-5} 
                        ~ & GADT-TIM & \textbf{91.3}    & \textbf{59.3} & \textbf{95.5} \\ \hline
\end{tabular}}
\caption{Attack success rate against defense models. With our proposed GADT strategy, the baseline, such as TIM, can reach a new peak, beating all current SOTA methods.}
\label{tab:defense}
\vspace{-0.1in}
\end{table}

\begin{figure}[t]
    \centering
    \includegraphics[width=\columnwidth]{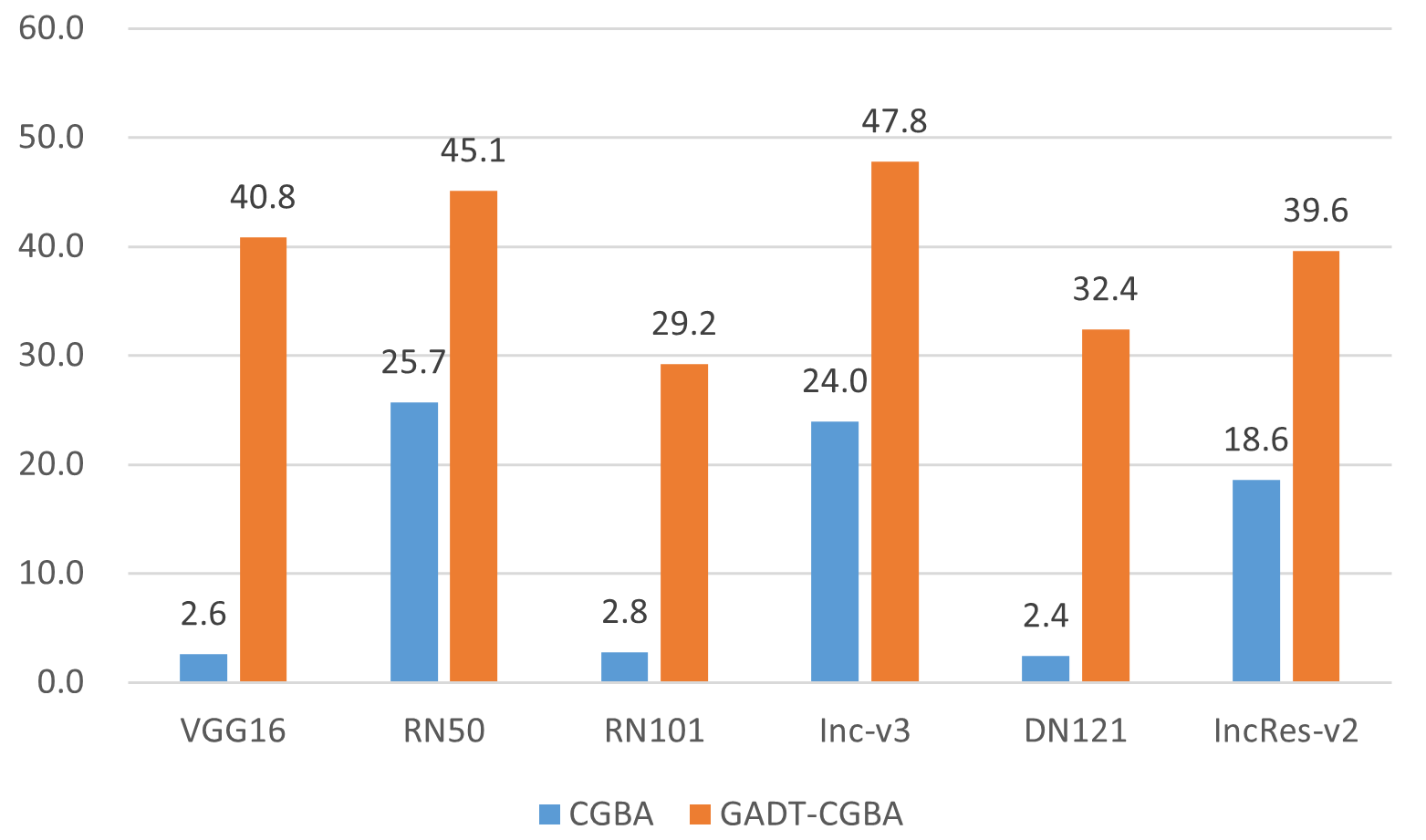}
    \caption{
    Tranferable attack success rate when GADT is combined with black box attack.
    }
    \label{figs:black}
    \vspace{-0.2in}
\end{figure}

Furthermore, our GADT method is a two-stage framework that synergizes effectively with existing attack methods, including those beyond transferable attacks. To demonstrate this capability, we integrated it with the query-based black-box attack CGBA~\cite{Reza_2023_ICCV}, evaluating the potential improvements in black-box attack success rates. Figure~\ref{figs:black} illustrates these results with RN50 as the target model, clearly showcasing the effectiveness of our approach. Particularly noteworthy is that our method improves the attack success rate by more than two times. 


\begin{table*}[h!]
    \centering
    \small
    \vspace{-0.1in}
    \begin{subtable}[t]{1.0\linewidth}
        \centering
        \renewcommand{\arraystretch}{1.05}
       \setlength{\tabcolsep}{0.5mm}{\begin{tabular}{c|c|c|c|c|c|c|c|c|c}
        \hline
            Surrogate & Attack & VGG16 & \makebox[0.06\textwidth][c]{RN50} & \makebox[0.06\textwidth][c]{RN101} & \makebox[0.06\textwidth][c]{Inc-v3} & DN121 & IncRes-v2 & $\rm CLIP_{RN101}$ & $\rm CLIP_{ViT/B32}$ \\ \hline 
            \multirow{2}{*}{VGG16} & MIM-k & 99.5 & 86.7 & 78.2 & 75.1 & 86.8 & 66.5 & 84.5 & 49.2 \\ 
                                 ~ & GADT-MIM   & \textbf{99.7} & \textbf{93.3} & \textbf{85.3} & \textbf{82.6} & \textbf{90.6} & \textbf{74.2} & \textbf{89.7} & \textbf{52.5} \\
            \hline
            \multirow{2}{*}{RN101} & MIM-k & 92.8 & 98.5 & 98.9 & 72.5 & 91.8 & 64.0 & 76.5 & 52.8 \\
                                 ~ & GADT-MIM   & \textbf{96.2} & \textbf{98.6} & \textbf{98.9} & \textbf{80.0} & \textbf{96.5} & \textbf{73.0} & \textbf{84.5} & \textbf{63.2} \\
            \hline
            \multirow{2}{*}{Inc-v3} & MIM-k & 86.6  & 72.8  & 63.4  & 99.3  & 70.0  & 69.0  & 66.2 & 38.6 \\ 
                                  ~ & GADT-MIM & \textbf{89.2}  & \textbf{79.0}  & \textbf{70.3}  & \textbf{99.0}  & \textbf{76.7}  & \textbf{77.1}  & \textbf{70.6} & \textbf{42.0} \\
            \hline
            \multirow{2}{*}{DN121} & MIM-k & 96.1  & 96.6  & 92.6  & 83.8  & 98.2  & 75.5  & 82.8 & 57.6 \\ 
                                 ~ & GADT-MIM & \textbf{98.4}  & \textbf{98.0}  & \textbf{95.4}  & \textbf{89.9}  & \textbf{98.8}  & \textbf{82.1}  & \textbf{88.0} & \textbf{72.1}\\ 
            \hline
        \end{tabular}}
    \end{subtable}
    \vspace{-0.05in}
    \caption{
    Ablation experiments: the comparisons with and without optimizing DA parameters.
    }
    \label{tab:ablation}
    \vspace{0.05in}
\end{table*}

\renewcommand{\dblfloatpagefraction}{0.7}
\begin{table*}[h!]
    \centering
    \small
    \begin{subtable}[t]{1.0\linewidth}
        \centering
        \setlength{\tabcolsep}{0.5mm}{\begin{tabular}{c|c|c|c|c|c|c|c|c|c}
        \hline
        Surrogate & Attack & VGG16 & \makebox[0.06\textwidth][c]{RN50} & \makebox[0.06\textwidth][c]{RN101} & \makebox[0.06\textwidth][c]{Inc-v3} & DN121 & IncRes-v2 & $\rm CLIP_{RN101}$ & $\rm CLIP_{ViT/B32}$ \\ \hline
        \multirow{3}{*}{VGG16}  & SIM-10 & 92.9  & 55.9  & 43.5  & 46.5  & 50.1  & 33.5  & 54.5 & 54.5 \\
               & SIM-30 & 92.6 & 53.9 & 42.6 & 42.0 & 47.4 & 30.5 & 53.6 & 58.5 \\
               & GADT-SIM   & \textbf{98.3} & \textbf{81.6} & \textbf{67.9} & \textbf{69.7} & \textbf{77.8} & \textbf{60.2} & \textbf{76.3} & \textbf{46.6} \\
               \hline
        \multirow{3}{*}{RN101}  & SIM-10 & 71.4  & 69.2  & 86.0  & 51.5  & 60.2  & 38.7  & 51.8 & 61.1 \\
               & SIM-30 & 70.4 & 63.7 & 78.3 & 40.4 & 50.4 & 29.6 & 47.0 & 66.0 \\
               & GADT-SIM   & \textbf{93.1} & \textbf{89.6} & \textbf{92.1} & \textbf{73.4} & \textbf{82.8} & \textbf{65.7} & \textbf{73.3} & \textbf{52.7} \\
               \hline
        \multirow{3}{*}{Inc-v3} & SIM-10 & 50.5  & 38.6  & 33.0  & 70.4  & 37.6  & 28.3  & 36.5 & 46.5 \\
               & SIM-30 & 45.6 & 32.4 & 28.2 & 61.4 & 30.6 & 22.8 & 34.4 & 46.4 \\
               & GADT-SIM   & \textbf{73.6} & \textbf{60.6} & \textbf{53.3} & \textbf{80.0} & \textbf{59.9} & \textbf{48.1} & \textbf{52.5} & \textbf{36.4} \\
               \hline
        \multirow{3}{*}{DN121}  & SIM-10 & 74.4  & 62.3  & 55.2  & 47.2  & 86.1  & 35.7  & 54.6 & 63.8 \\
               & SIM-30 & 71.9 & 58.6 & 50.4 & 43.7 & 80.5 & 31.3 & 53.7 & 67.4 \\
               & GADT-SIM   & \textbf{93.0} & \textbf{85.1} & \textbf{78.7} & \textbf{70.5} & \textbf{94.4} & \textbf{63.5} & \textbf{77.8} & \textbf{54.6} \\
               \hline
        \end{tabular}}
    \end{subtable}
    \vspace{-0.05in}
    \caption{
    Ablation experiments: comparing GADT with the attack baselines with varying iteration numbers.
    }
    \label{tab:iteration}
    \vspace{0.05in}
\end{table*}

\subsection{Attacks for Models with Defense Mechanism}
In this section, we combine our method with TIM to attack adversarially trained Inception-v3~\citep{kurakin2018adversarialattacksdefencescompetition}, the ensemble adversarial training strategy (AT)~\citep{tramer2017ensemble}, and the defense strategy of HGD~\citep{liao2018defense}, evaluating the effectiveness of our attack against various defense techniques. The results are shown in Table~\ref{tab:defense}. Based on the experimental results, we observe that our method significantly enhances the attack effectiveness of the weak baseline (TIM), consistently improving performance across various target models. Compared to existing methods, GADT-TIM commonly achieves a higher attack success rate. GADT-TIM even outperformed over by 20\% on HGD when the surrogate model is DN121. Especially in attacks against $\rm Inc\text{-}v3_{adv}$ and AT, we have achieved over a 10\% increase in attack success rate. In summary, when combined with GADT, attack baselines, such as TIM, can outperform the current SOTA DA-based method, e.g., Admix. Moreover, higher attack effectiveness can be achieved by combining GADT with stronger baseline methods.

\subsection{Ablation Study}

\noindent\textbf{The effectiveness of our DA optimization strategy.}
To validate the effectiveness of our DA optimization strategy, we conducted ablation experiments by removing our gradient-guided DA optimization procedure. Similar to our full strategy, we combined the same Kornia-based DA transformation operations with MIM but without the DA optimization process, denoted as MIM-k. The results are listed in Table~\ref{tab:ablation}. We observed that attacks using our original GADT strategy generally outperformed those without the corresponding DA optimization. Specifically, when attacking IncRes-v2, the attack success rate increased by 10\% when comparing GADT-MIM and MIM-k. Moreover, for attacks on $\rm CLIP_{RN101}$ and $\rm CLIP_{ViT-B/32}$, there was an improvement of over 4\%. Experiments involving other attack methods also support a similar conclusion, as shown in Appendix B.
These experimental results underscore the necessity of optimizing DA operations with our strategy. Our approach effectively expands the solution space for generating adversarial samples against various target models.

\noindent\textbf{Is the advantage of our method solely due to the additional iterations in the first stage?}
Compared with the baseline, our method involves additional attack iterations in the first stage, perturbing and optimizing the DA parameters. Some may doubt whether our superiority is mainly due to these additional iterations. To address this, we establish a comparison baseline: the attack baseline with the same number of iterations as GADT-X, named ``Y-Z", where Y represents the attack method's name and ``Z" the iteration count. In our previous experiments, ``Z=10", and our DA optimization iteration number is 20. Therefore, we set ``Z=30" in the experiments of this section, ensuring that our method and ``Y-30" have the same iteration count. As shown in Table~\ref{tab:iteration}, comparing ``Y-10" with ``Y-30", it is evident that increasing attack iterations improves the attack success rate but within a limited range. However, our method, GADT-X, consistently outperforms the baseline across varying iterations, with at least a 5\% improvement compared to all ``Y-30" settings. More results can be seen in Appendix C.


\noindent\textbf{The effect of our loss function $\mathcal{L}_{trans}$ on fidelity.}
Our method specifically enhances the fidelity of adversarial examples by designing a loss function, $\mathcal{L}_{trans}$. To verify its effectiveness, we selected two image quality assessment metrics: PSNR and SSIM, to assess the similarity between adversarial examples generated by MIM/GADT-MIM and clean examples. We computed the average values of these metrics for the entire dataset. Table~\ref{tab:simi} lists the results, showing that higher scores indicate greater similarity. Regardless of the surrogate model used, our method consistently achieves higher scores and fidelity, demonstrating the effectiveness of $\mathcal{L}_{trans}$ for the fidelity.

\begin{table}[t]
\vspace{-0.1in}
\small
\renewcommand{\arraystretch}{1.3}
\scalebox{0.9}{\begin{tabular}{c|cc|cc}
\hline
\multirow{2}{*}{Surrogate} & \multicolumn{2}{c|}{PSNR}                         & \multicolumn{2}{c}{SSIM}                          \\ \cline{2-5}  
                           & \multicolumn{1}{c|}{MIM}      & GADT-MIM           & \multicolumn{1}{c|}{MIM}      & GADT-MIM           \\ \hline  
VGG16                      & \multicolumn{1}{c|}{12.62926} & \textbf{12.78718} & \multicolumn{1}{c|}{0.09293} & \textbf{0.09545} \\ \hline
RN101                      & \multicolumn{1}{c|}{12.64113} & \textbf{12.80395} & \multicolumn{1}{c|}{0.09428} & \textbf{0.09687} \\ \hline
DN121                      & \multicolumn{1}{c|}{12.64379}  & \textbf{12.80329} & \multicolumn{1}{c|}{0.09433} & \textbf{0.09686} \\ \hline
\end{tabular}}
\caption{The comparisons with baselines in terms of the fidelity between adversarial and clean samples.}
\label{tab:simi}
\vspace{-0.1in}
\end{table}

\section{Conclusion}
In this paper, we introduce a novel DA optimization strategy aimed at generating effective and transferable adversarial examples, termed GADT. Unlike existing approaches, we compute gradients of the loss with respect to DA parameters, leveraging the differentiable DA operations provided by Kornia. Additionally, we design a new loss function to guide the optimization of DA parameters, balancing attack effectiveness and stealthiness. Our approach is compatible with all existing transferable attack strategies, and extensive experiments validate the improvements achieved by incorporating GADT. GADT can be extended to other black-box attack strategies, offering new insights for attack algorithms.


\bibliography{aaai25}

\end{document}